\documentclass{bmvc2k}
\usepackage[export]{adjustbox}
\usepackage{times}
\usepackage{epsfig}
\usepackage{graphicx}
\usepackage{amsmath}
\usepackage{amssymb}
\usepackage[export]{adjustbox}

\usepackage{dsfont} 

\usepackage{graphics}
\usepackage{array}
\usepackage{multirow}
\usepackage{dblfloatfix}

\usepackage{wrapfig}

\usepackage{lipsum}

\usepackage{comment}
\usepackage{mathtools}
\usepackage[]{algorithm2e}

\usepackage{dblfloatfix}

\usepackage{paralist}
\usepackage{enumerate}
\usepackage{enumitem}

\usepackage[export]{adjustbox}

\usepackage{comment}
\usepackage{dsfont} 
\usepackage{color}
\usepackage{pifont}

\usepackage{multirow}

\usepackage{xcolor}

\usepackage[utf8]{inputenc}
\usepackage{float}

\usepackage{booktabs}
\usepackage{array}
\usepackage{multirow}
\usepackage{dblfloatfix}
\usepackage{enumitem}

\usepackage{pifont}% http://ctan.org/pkg/pifont
\usepackage{sidecap}

\def\eg{\emph{e.g}.} 
\def\Eg{\emph{E.g}.}
\def\etc{\emph{etc}.}

\def\ie{\emph{i.e}.}

\def\wrt{w.r.t.}
\DeclareMathOperator*{\argmax}{arg\,max}

\title{Structured Latent Embeddings for \\ Recognizing Unseen Classes in Unseen Domains}

\addauthor{Shivam Chandhok}{shivam.chandhok@mbzuai.ac.ae}{1}
\addauthor{Sanath Narayan}{sanath.narayan@inceptioniai.org}{2}
\addauthor{Hisham Cholakkal}{hisham.cholakkal@mbzuai.ac.ae}{1}
\addauthor{Rao Muhammad Anwer}{rao.anwer@mbzuai.ac.ae}{1}
\addauthor{Vineeth N Balasubramanian}{vineethnb@iith.ac.in}{4}
\addauthor{Fahad Shahbaz Khan}{fahad.khan@mbzuai.ac.ae}{13}
\addauthor{Ling Shao}{ling.shao@ieee.org}{2}
\addinstitution{
Mohamed Bin Zayed University \\of AI, UAE
}
\addinstitution{
 Inception Institute of Artificial\\ Intelligence, UAE
}
\addinstitution{
Linkoping University, Sweden}
\addinstitution{
Indian Institute of Technology,\\
Hyderabad, India}
\runninghead{Shivam et al.}{Recognizing Unseen Classes in Unseen Domains}

\def\eg{\emph{e.g}\bmvaOneDot}
\def\Eg{\emph{E.g}\bmvaOneDot}

\begin{document}

\maketitle

\begin{abstract}
The need to address the scarcity of task-specific annotated data has resulted in concerted efforts in recent years for specific settings such as zero-shot learning (ZSL) and domain generalization (DG), to separately address the issues of semantic shift and domain shift, respectively. However, real-world applications often do not have constrained settings and necessitate handling unseen classes in unseen domains -- a setting called Zero-shot Domain Generalization, which presents the issues of domain and semantic shifts simultaneously. In this work, we propose a novel approach that learns domain-agnostic structured latent embeddings by projecting images from different domains as well as class-specific semantic text-based representations to a common latent space. In particular, our method  jointly strives for the following objectives: 
(i) aligning the multimodal cues from visual and text-based semantic concepts; (ii) partitioning the common latent space according to the domain-agnostic class-level semantic concepts; and (iii) learning a domain invariance \wrt{} the visual-semantic joint distribution for generalizing to unseen classes in unseen domains. Our experiments on the challenging DomainNet and DomainNet-LS benchmarks show the superiority of our approach over existing methods, with significant gains on difficult domains like \textit{quickdraw} and \textit{sketch}.
\end{abstract}

\section{Introduction}
\label{intro}
\vspace{-2pt}
In various  computer vision  problems, obtaining labeled data specifically tailored for a new task (be it a new domain or a new class) can be challenging due to one or more of several reasons: high annotation costs, dynamic addition of objects with new semantic content, limited instances of rare objects or long-tailed distributions which frequently occur in real-world scenarios \cite{dascn}. To address such issues, two popular recent approaches include: (i) zero-shot learning (ZSL): use training data of  related  object categories from the same domain (\eg, sketches of cats as a training data for recognizing dogs from sketches); and (ii) domain generalization (DG): use training data of a particular object category from related domains (\eg, photos/real images of dogs as a training data for recognizing dogs from sketches). More recently, there has been increasing interest in a combination of these approaches to handle unseen classes in unseen domains, viz. zero-shot domain generalization (which we call ZSLDG), where one leverages training data of a related object category from a related domain (\eg, photos of cats as a training data for recognizing dogs from sketches). DG addresses only domain shift that occurs due to the training and test domains being different. ZSL addresses semantic shift that occurs due to the presence of different object categories during training and testing. In contrast, ZSLDG more closely aligns with the challenges faced in real-world applications, but needs to simultaneously address domain and semantic shift issues \cite{cumix,ZeroShotDG}. 
\begin{wrapfigure}{R}{0.5\columnwidth}
    \centering
    \includegraphics[width=0.48\textwidth]{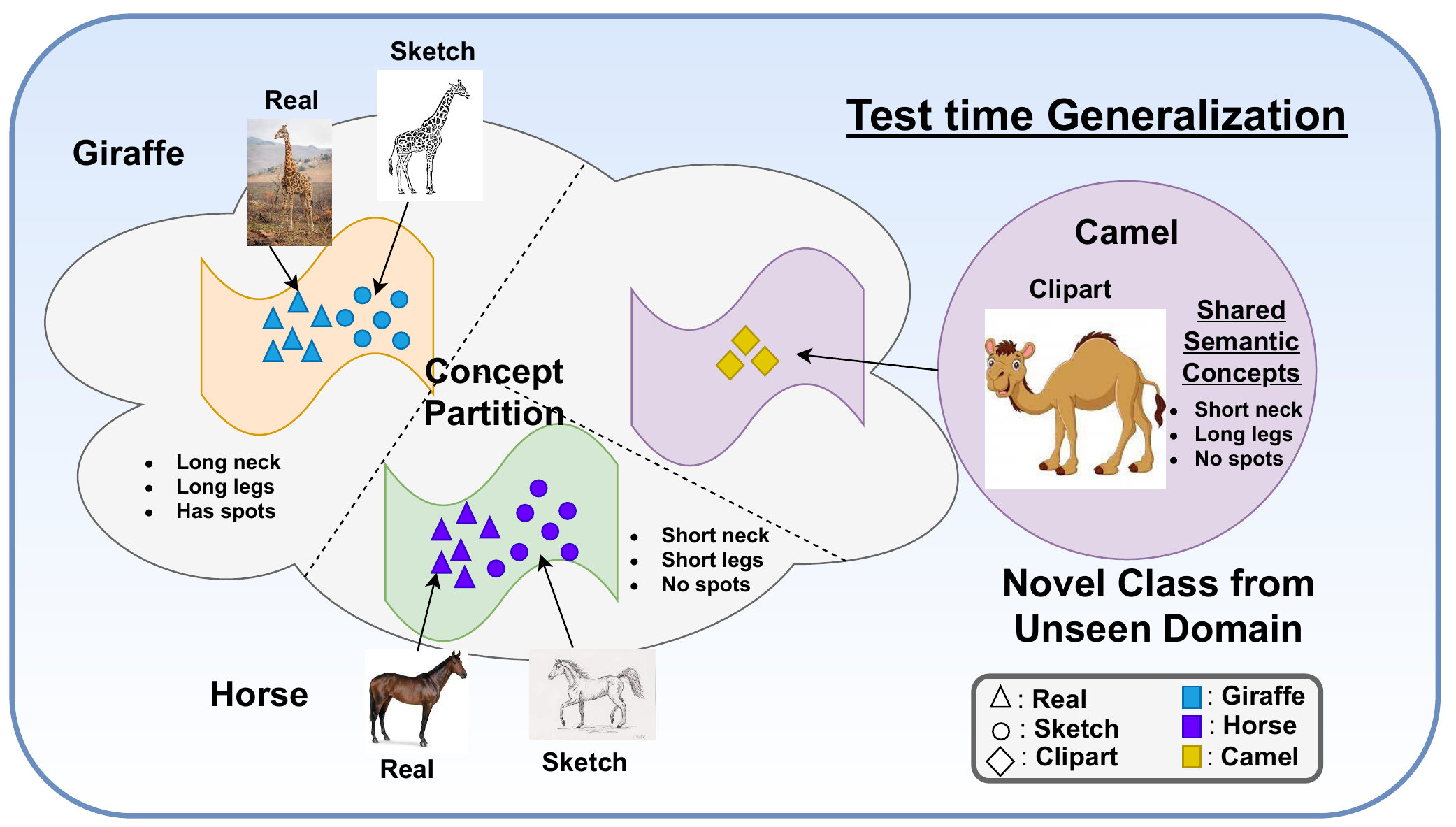}\vspace{0.15cm}
    \caption{\footnotesize \label{fig:intro_fig}Our latent space is structured according to class-level semantic concepts and is domain-invariant \wrt{} visual-semantic joint distribution. This enables our model to map unseen classes in unseen domains at test-time (\textit{camel} from \textit{clipart}), based on their semantic concepts, to appropriate subspaces within our latent space, thereby aiding generalization.}
\vspace{-8pt}
\end{wrapfigure}
% At test time, when our model encounters an image from an unseen domain (\textit{giraffe} from \textit{sketch} domain), it recognizes the class using the visual-semantic relationships learned during training.
In this work, we investigate this challenging problem of ZSLDG. In particular, we propose a unified solution that jointly tackles both domain and semantic shifts by relating the general visual cues of a class to semantic concepts that are invariant across domains. \Eg, semantic cues such as  <\textit{long neck}, \textit{long legs}, \textit{has spots}> of \textit{giraffe} class are invariant across domains such as \textit{real} images (photos), \textit{sketch} or \textit{clipart} (see Fig.~\ref{fig:intro_fig}). To this end, we bring common information from visual and semantic spaces into a structured domain-agnostic latent space, partitioned according to class-level semantic concepts. We then impose a domain invariance \wrt{} the visual-semantic joint distribution. Since the semantic space is shared across all classes and is agnostic to the visual domains, imposing such invariance aids in generalizing to unseen domains at test time (instead of overfitting to source domains), while improving the visual-semantic interaction for effective knowledge transfer across seen and unseen classes.

\noindent\textbf{Contributions:}
We propose a ZSLDG approach comprising of
%that comprises 
a visual encoder that learns to project multi-domain images from the visual space to a latent space, and a semantic encoder that learns to map text-based category-specific semantic representations to the same latent space.
The key contributions of the proposed approach are: 
\textbf{(i)} For aligning class-specific cues from visual and semantic latent embeddings, we introduce a multimodal alignment loss term; 
%(ii) We propose to obtain a domain-agnostic class-level partitioning of the latent space, we introduce a center loss term that minimizes the intra-class variance across different seen domains
\textbf{(ii)} We propose to partition the latent space \wrt{} class-level semantic concepts across domains by minimizing intra-class variance across different seen domains; % through a center loss term;
%\textbf{(iii)} To further achieve a domain invariance \wrt{} the visual-semantic joint distribution, we introduce a joint invariance module that facilitates generalizing to unseen classes in unseen domains;
\textbf{(iii)} The focus of our design is introduction of a joint invariance module that seeks to achieve domain invariance \wrt{} the visual-semantic joint distribution, and thereby facilitates generalizing to unseen classes in unseen domains; % Our  joint invariance module 
%To further achieve a domain invariance \wrt{} the visual-semantic joint distribution, we introduce a joint invariance module that facilitates generalizing to unseen classes in unseen domains;
and \textbf{(iv)} Experiments and ablation studies on the challenging DomainNet and DomainNet-LS benchmarks~\cite{domainnet} demonstrate the superiority of our approach over existing methods. Particularly, on the most difficult \textit{quickdraw} domain, our approach achieves a significant gain of $1.6\%$ over the best existing method~\cite{cumix}.

\section{Related Work\label{related}}
\noindent\textbf{Domain Generalization (DG):}  Existing methods tackle the problem of domain shift, which occurs when the training and testing data belong to different domains, in different ways. Most previous approaches aim to learn domain-invariance by minimizing the discrepancy between multiple source domains\cite{basicdg1,basicdg2,basicdg3} or by employing autoencoders and adversarial losses \cite{MTAE,DAFL}. A few works~\cite{metadg1,metadg2,metadg3} introduce specific training policies or optimization procedures such as meta-learning and episodic training to enhance the generalizability of the model to unseen domains. Similarly, \cite{crossgrad,advaug} employ data augmentation strategies to improve the models robustness to data distribution shifts at test time. However, all these works tackle the DG problem alone, where the label spaces at both train and test time are identical.\\
\noindent\textbf{Zero-shot Learning (ZSL):} Traditional ZSL methods~\cite{zslthree,zslfour,zslfive,zslsix,zslseven} learn to project the visual features onto a semantic embedding space via direct mapping or through a compatibility function. However, such direct mappings are likely to suffer from issues of seen class bias and hubness~\cite{zsltwentyfour,zsltwentyfive}. In contrast, the work of~\cite{zsltwelve} leverages joint multi-modal learning of visual and textual feature embeddings for the task of ZSL. Recently, generative approaches tackle the problem of seen class bias by generating unseen visual features from respective class embeddings\cite{zslfourteen,zslsixteen,lisgan,lsrgan,zslfifteen,zslthirteen,vaegan,tfvaegan}. However, all the aforementioned methods address only ZSL, where the domain remains unchanged during training and testing. \\
\noindent\textbf{Zero-shot Domain Generalization (ZSLDG):} 
%Though considerable research has been undertaken in the standard DG and ZSL settings, the problem of ZSLDG was only recently introduced in CuMix~\cite{cumix}. 
Recently, CuMix~\cite{cumix} introduced the problem of ZSLDG. 
While~\cite{ZeroShotDG} defined variations in rotations of the same objects as different domains, such a restricted definition limits its real-world applicability. Differently, CuMix~\cite{cumix} defines domains as different ways of depicting an object, as in \textit{sketch}, \textit{painting}, \textit{cartoon}, \etc, which is closer to practical use of such methods.
% The work of~\cite{ZeroShotDG}   defines domains as the variations in rotations of the same object, whereas CuMix~\cite{cumix} defines domains as different ways of depicting an object, as in \textit{sketch}, \textit{painting}, \textit{cartoon}, \etc{} 
% In this work, we define domains as in CuMix~\cite{cumix} since it has large number of real-world applications. 
%  as different domains,  such a restricted definition limits its real-world applicability. Differently,
CuMix tackles the issue of domain shift through data augmentation by mixing and interpolating source domains, and handles semantic shifts by learning to project visual features to the semantic space. 
This work also established a benchmark dataset, DomainNet, for this setting with an evaluation protocol, which we follow in this work for fair comparison. 
However, relying on mixing source domains has a drawback -- the resulting model could overfit to the source domains and their interpolations, thereby reducing generalizability to unseen domains~\cite{dgoverfit}.
%, as investigated in the DG approach~\cite{dgoverfit}.
%----original text
% \noindent\textbf{Zero-shot Domain Generalization (ZSLDG):} Though considerable research has been undertaken in the standard DG and ZSL settings, the problem of ZSLDG was only recently introduced in CuMix~\cite{cumix}. While the work of~\cite{ZeroShotDG} defines the variations in rotations of the same objects as different domains, such a restricted definition limits its real-world applicability. Differently, CuMix~\cite{cumix} defines domains as different ways of depicting an object, as in \textit{sketch}, \textit{painting}, \textit{cartoon}, \etc. Furthermore, it tackles the issue of domain shift through data augmentation by mixing and interpolating the source domains, while handling the semantic shifts by learning to project the visual features to the semantic space. 
% % This work also established a benchmark dataset, DomainNet, for this setting with an evaluation protocol, which we follow in this work for a fair comparison. 
% However, mixing source domains is likely to result in overfitting to the source domains and their interpolations, thereby reducing generalizability to unseen domains, as investigated in the DG approach~\cite{dgoverfit}.
% However, mixing source domains is less likely to generalize to unseen domains since the learned representation tends to overfit to the source domain distribution and their interpolations, as has been previously found in~\cite{dgoverfit}, which addresses only DG. 
Furthermore, directly mapping the visual space to the semantic space, as in~\cite{cumix}, can lead to hubness issues (mapped points cluster as a hub due to low variance)~\cite{zsltwentyfour,zsltwentyfive}, thereby reducing class-discriminative capability. In contrast, our approach jointly handles the issues of domain and semantic shifts by  learning a domain-agnostic latent space that is partitioned based on class-level (domain-invariant) semantic concepts, onto which the visual and semantic features are projected. Since domain invariance is enforced \wrt{} the visual-semantic joint distribution, it is less likely to overfit to seen domains (a common problem when domain-invariance is enforced \wrt{} marginal distribution of images~\cite{dgoverfit,condinvariant}). In addition, our approach enables better interaction between visual and semantic spaces in a new latent space, thereby supporting model generalization to unseen classes in unseen domains.

\vspace{-7pt}
\section{Proposed Method\label{model}}

\noindent \textbf{Problem Setting:} The goal in zero-shot domain generalization (ZSLDG) is to recognize unseen categories in unseen domains. Let  $Q^{Tr}=\{ (\mathbf{x},  y,  \mathbf{a}_y, d )| \mathbf{x} \in \mathcal{X},  y \in \mathcal{Y}^s,  \mathbf{a}_y \in \mathcal{A},  d \in \mathcal{D}^s\}$ denote the training set, where $\mathbf{x}$ is a seen class image in the visual space ($\mathcal{X}$) with corresponding label $y$ from a set of seen class labels $\mathcal{Y}^s$. Here, $\mathbf{a}_y$ denotes the class-specific semantic representation that encodes the inter-class relationships, while $d$ is the domain label from a set of seen domains $\mathcal{D}^s$. Note that the semantic representations are typically obtained from unsupervised text-based WordNet models (\eg, \textit{word2vec}~\cite{word2vec})
Similarly, $Q^{Ts}=\{ (\mathbf{x},  y,  \mathbf{a}_y, d )| \mathbf{x} \in \mathcal{X},  y \in \mathcal{Y}^u,  \mathbf{a}_y \in \mathcal{A},  d \in \mathcal{D}^u\}$ is the test set, where $\mathcal{Y}^u$ is the set of labels for unseen classes and $\mathcal{D}^u$ represents the set of unseen domains. In the standard zero-shot setting, images at training and testing belong to disjoint classes but share the same domain space, \ie, $\mathcal{Y}^s\cap\mathcal{Y}^u \equiv \emptyset$ and $\mathcal{D}^s\equiv\mathcal{D}^u$. On the other hand, in the standard DG setting, images at training and testing belong to same categories in disjoint domain spaces, \ie, $\mathcal{Y}^s\equiv\mathcal{Y}^u $ and $\mathcal{D}^s \cap \mathcal{D}^u \equiv \emptyset$. 
% Note that this implies that the conditional distribution $p(y|\mathbf{x})$ changes within the training set as well since $\mathbf{x}$ comes from different domains, \ie, $p_\mathcal{X}(\mathbf{x}|d_i)\neq p_\mathcal{X}(\mathbf{x}|d_j)$, $\forall i\neq j$.
In this work, our goal is to address the more challenging ZSLDG setting for recognizing unseen classes in unseen domains without having seen these novel classes and domains during training, \ie, $\mathcal{Y}^s\cap\mathcal{Y}^u \equiv \emptyset$ and $\mathcal{D}^s\cap\mathcal{D}^u \equiv \emptyset$.

\begin{figure*}[t]
    \centering
    \includegraphics[width=0.95\textwidth]{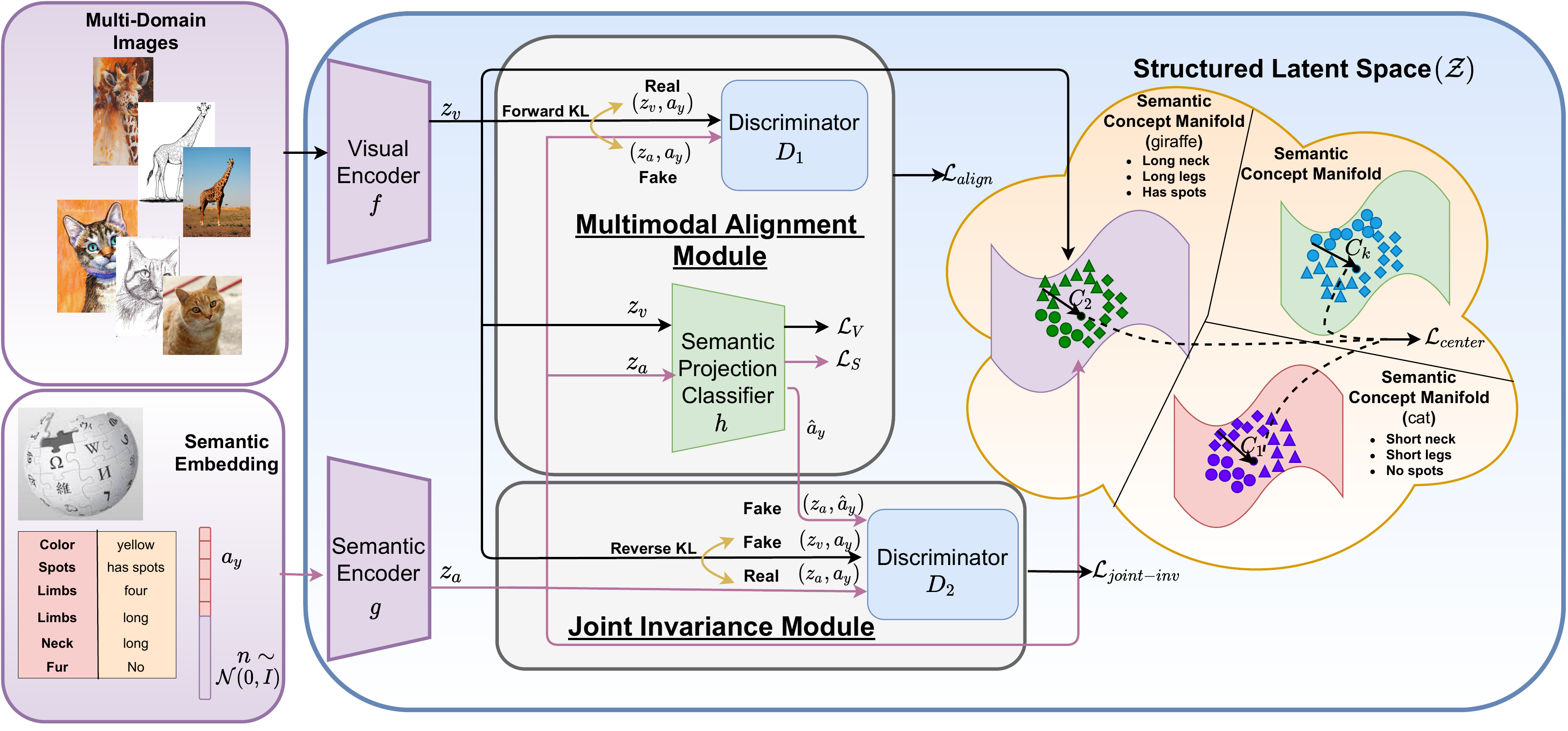}\vspace{0.15cm}
    \caption{\footnotesize Overall architecture of our approach. The proposed approach comprises a visual encoder $f$ and a semantic encoder $g$. The multimodal alignment module (Sec.~\ref{alignment}) aligns the class-specific cues from the visual and semantic latent embeddings ($\mathbf{z}_{v}$ and $\mathbf{z}_{a}$) in $\mathcal{Z}$ by employing an alignment loss term $\mathcal{L}_{align}$, The loss term $\mathcal{L}_{center}$ ensures a domain-agnostic class-level partitioning (Sec.~\ref{struct_embed}) of $\mathcal{Z}$. Furthermore, the joint invariance module strives to achieve domain invariance (Sec.~\ref{triplegan}) \wrt{} the visual-semantic joint distribution by employing $\mathcal{L}_{joint-inv}$, thereby enabling us to generalize to unseen classes in unseen domains.}
    \label{fig:main_fig}
    \vspace{-0.2cm}
\end{figure*}

\noindent\textbf{Overall Framework:} 
% \subsection{Overall Framework}
% \textcolor{red}{Here, we describe the ideology and motivation behind our approach and then introduce various components in our proposed framework for recognizing unseen classes in unseen domains.}  
The overall architecture of our proposed approach is shown in Fig.~\ref{fig:main_fig}. The proposed framework comprises a visual encoder $f$, semantic encoder $g$, semantic projection classifier $h$ along with discriminators $D_{1}$ and $D_{2}$. In ZSLDG, the conditional distribution $p(y|\mathbf{x})$ changes since $\mathbf{x}$ comes from different domains, \ie, $p_\mathcal{X}(\mathbf{x}|d_i)\neq p_\mathcal{X}(\mathbf{x}|d_j)$, $\forall i\neq j$. Our approach mitigates this issue by learning a domain-invariant semantic manifold $\mathcal{Z}$ which is partitioned according to class-level semantic concepts (described in Sec.~\ref{alignment} and Sec.~\ref{struct_embed}), such that $p(y|\mathbf{z})$ is stable and does not change across domains (where $\mathbf{z} = f(\mathbf{x}$)). Furthermore, in order to ensure generalization to unseen classes in unseen domains at test time, our joint invariance module achieves domain invariance \wrt{} the visual-semantic joint by employing $\mathcal{L}_{joint-inv}$ (described in Sec.~\ref{triplegan}). This facilitates improved knowledge transfer between  class-specific (domain-invariant) visual cues and semantic representations in latent space $\mathcal{Z}$, thereby enhancing generalization to unseen classes in unseen domains at test-time.

\vspace{-7pt}
\subsection{Multimodal Alignment\label{alignment}}
%As discussed earlier, our aim is to learn a domain invariant latent embedding space which is able to recognize different novel classes across domains. In order to recognize novel classes, latent space need to be partitioned/structured according to the inter-class relations. To achieve this goal, we first learn a joint latent space between the visual and semantic modalities in our proposed multimodal alignment module. Our multimodal alignment module learns to project both the visual and semantic representations to a joint latern embedding space.
 The multimodal alignment module, learns to project both the visual and semantic representations to a common latent embedding space $\mathcal{Z}$. 
% 
% The objective of our multimodal alignment module is to learn a joint conceptual latent space, where the common information across domains from visual modality and semantic information space can be accumulated and related. 
% Furthermore, aligning distributions in a common latent space is also found to facilitate knowledge transfer between visual and semantic spaces. 
% 
Let $f(\mathbf{x}):\mathcal{X}\rightarrow \mathcal{Z}$ denote a feature extractor, which maps an image $\mathbf{x}$ in the visual space $\mathcal{X}$ to a vector $\mathbf{z}_{v}$ in the latent embedding space $\mathcal{Z}$. Furthermore, let the function $g$ learn a mapping from semantic space to the latent embedding space, \ie, $g(\mathbf{n},\mathbf{a}_y):\mathcal{N}\times\mathcal{A}\rightarrow \mathcal{Z}$ by taking a random Gaussian noise vector $\mathbf{n}$ concatenated with the semantic representation $\mathbf{a}_y$ as input and mapping it to a vector $\mathbf{z}_{a}$ in $\mathcal{Z}$. Let $D_1: \mathcal{Z} \times \mathcal{A} \rightarrow \mathbb{R}$ denote a conditional discriminator (conditioned on the semantic embedding $\mathbf{a}_y$) . Then, the multimodal adversarial alignment of the visual and semantic embedding spaces is achieved by employing a Wasserstein GAN~\cite{wgan}, as given by
\begin{equation}
\label{eq:wganD1}
\mathcal{L}_{D_{1}} = \hspace{0.2em} \mathbb E[D_1(\mathbf{z}_{v}, \mathbf{a}_y)] - \mathbb E[D_1({\mathbf{z}_{a}}, \mathbf{a}_y)] -
                     \lambda \mathbb E[\left(||\nabla_{\tilde{\mathbf{z}}} D_1(\tilde{\mathbf{z}}, \mathbf{a}_y)||_2 - 1\right)^2], 
% \vspace{-3pt}      
\end{equation}

\noindent where ${\mathbf{z}_{v}}=f(\mathbf{x})$ and ${\mathbf{z}_{a}}=g(\mathbf{n},\mathbf{a}_y)$ are the latent embeddings from the visual and semantic spaces, respectively. Here, $\lambda$ is a weighting coefficient, while $\tilde{\mathbf{z}} = \eta \mathbf{z}_{v} +(1-\eta){\mathbf{z}_{a}}$ with $\eta \sim U(0, 1)$ represents a convex combination of $\mathbf{z}_v$ and $\mathbf{z}_a$. Eq.~\ref{eq:wganD1} is equivalent to minimizing the (forward) Kullback-Leibler (KL) divergence between the visual and semantic latent embeddings, \ie, $KL[(\mathbf{z}_{v},\mathbf{a}_{y})||(\mathbf{z}_{a},\mathbf{a}_{y})]$.
% Furthermore, in order to ensure discriminative conceptual signatures in the latent space, we impose a classification loss on the latent embeddings from visual and semantic spaces \emph{i.e} $\mathbf{z}_{v}$ and $\mathbf{z}_{a}$.
Furthermore, to enhance the discriminability of learned latent embeddings, we employ a compatibility based classifier using a semantic projection function $h:\mathcal{Z}\rightarrow \mathcal{A}$ for constraining the latent embeddings ($\mathbf{z}_v$ and $\mathbf{z}_a$) to map back to their corresponding semantic representations $\mathbf{a}_y$, given by, 
\begin{equation}
\label{eq:wgan2}
\mathcal{L}_{V}(\mathbf{z}_{v},\mathbf{a}_y)=-\mathbb E(\log\frac{\exp(\langle h(\mathbf{z}_{v}),\mathbf{a}_y\rangle)}{ \Sigma_{\mathbf{y}\in Y^{s}}\exp(\langle h(\mathbf{z}_{v}),\mathbf{a}_y\rangle)}),\hspace{5pt}
\mathcal{L}_{S}(\mathbf{z}_{a},\mathbf{a}_y)=-\mathbb E( \log\frac{\exp(\langle h(\mathbf{z}_{a}),\mathbf{a}_y\rangle)}{ \Sigma_{\mathbf{y}\in Y^{s}}\exp(\langle h(\mathbf{z}_{a}),\mathbf{a}_y\rangle)}).
 \vspace{-0.1cm}                 
\end{equation}
% 
% 
% 
% \begin{align}
% \label{eq:wgan3}
% \mathcal{L}_{S}(\mathbf{z}_{a},\mathbf{a}_y)=-\mathbb E( \log\frac{\exp(\langle h(\mathbf{z}_{a}),\mathbf{a}_y\rangle)}{ \Sigma_{\mathbf{y}\in Y^{s}}\exp(\langle h(\mathbf{z}_{a}),\mathbf{a}_y\rangle)}).
% % \vspace{-3pt}                 
% \end{align}
% 
% 
Here, $\langle\cdot,\cdot\rangle$ represents the measure of similarity between its inputs, computed as the dot product between them.
% This mainly ensures that the representation of a particular class in the latent is compatible with the semantic concepts of the class. 
Such a cyclic projection, \ie, mapping from visual/semantic space to a latent space and then back to the semantic space minimizes the information loss and enhances the discriminabilty of the latent embeddings. We employ the multimodal alignment loss term ($\mathcal{L}_{align}$) to learn the visual and semantic encoders along with the semantic projection classifier, given by
\begin{equation}
\label{eq:align}
\mathcal{L}_{align} =  \mathbb E[D_1(\mathbf{z}_{v}, \mathbf{a}_y)] - \mathbb E[D_1({\mathbf{z}_{a}}, \mathbf{a}_y)] + \mathcal{L}_{V}(\mathbf{z}_v,\mathbf{a}_y) + \mathcal{L}_{S}(\mathbf{z}_a,\mathbf{a}_y)].
% \vspace{-3pt}                 
\end{equation}

\vspace{-7pt}
\subsection{Structured Partitioning\label{struct_embed}}
\begin{wrapfigure}{R}{0.4\columnwidth}
    \centering
    \includegraphics[width=0.35\textwidth]{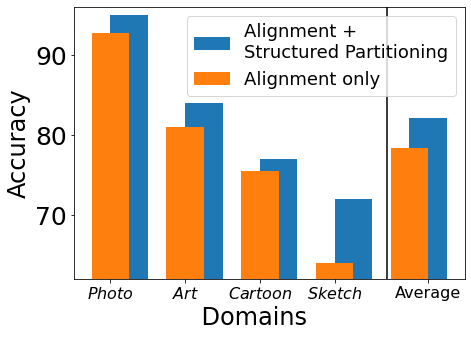}\vspace{0.1cm}
    \caption{\label{fig:dg}\footnotesize Impact of our structured partitioning for the DG task on PACS~\cite{basicdg4}. Compared to multimodal alignment alone (orange bars), additionally partitioning the latent space according to the semantic concepts along with multimodal alignment provides notable performance gains (blue bars), especially on the most difficult unseen domain, \ie, \textit{sketch}.}
    \vspace{-9pt}
\end{wrapfigure}
While the multimodal alignment aligns the visual and corresponding semantic embeddings in the latent space, it does not learn a domain-agnostic latent space, which is partitioned according to the semantic concepts that relate to the different classes.
% In addition, since novel classes need to be recognized, structuring/partitioning the latent space according to the inter-class relationships becomes necessary. 
% Aligning multi-domain visual features and semantic features does not ensure a structured latent space. 
% Learning a structured and domain-agnostic latent embedding space is desired for tackling the semantic shifts (novel classes) and domain shifts (novel domains) at test time. 
% While the multimodal alignment aligns the visual and corresponding semantic embeddings in the latent space, it does not learn a domain-agnostic latent space, which is structured according to the semantic concepts that relate to the different classes. 
% we want our latent space to be domain agnostic and at the same time partitioned based on conceptual semantic signatures.
% 
%
% 
% 
In order to achieve a structured and domain-invariant latent space, we propose to cluster the latent embeddings based on class-level (domain-invariant) semantic concepts  across different domains. The latent space is then conceptually structured, since the visual latent embeddings $\mathbf{z}_v$ and semantic latent embeddings $\mathbf{z}_a$ of a class are clustered together. 
% Since the visual features of a category across different domains are clustered together along with the corresponding semantic latent embeddings of their class, the resulting latent space is  domain agnostic and class-wise discriminative.
To this end, we adopt the center loss ~\cite{centerloss} in a multimodal setting. Formally, we first randomly initialise $S$ centers, \ie, $\{\mathbf{c}_{j}|j=1,\ldots, S\}$ for each of the seen classes in the training set and compute the loss, $\mathcal{L}_{center}$ due to each class $y$ present in a mini-batch. Then, for every class $y$ that is present in a mini-batch, the center update $\Delta \mathbf{c}_y$ is computed for incrementing the corresponding center $\mathbf{c}_y$. The loss $\mathcal{L}_{center}$ and update $\Delta \mathbf{c}_y$ are given by:
\begin{align}
\label{eq:center_loss}
\mathcal{L}_{center}=\delta[\mathbb E(||\mathbf{z}_{v}-\mathbf{c}_y||_{2}^2)+\mathbb E(||\mathbf{z}_{a}-\mathbf{c}_y||_{2}^2)]; \qquad & \Delta \mathbf{c}_y = \mathbb E [\mathbf{c}_y-\mathbf{z}_v] + \mathbb E [\mathbf{c}_y-\mathbf{z}_a].
% \vspace{-3pt}                 
\end{align}
Here, $\mathbf{c}_y$ denotes the center of class label $y$ in the latent space, while $\mathbf{z}_v$ and $\mathbf{z}_a$ correspond to the visual and semantic embeddings of class $y$, and $\delta$ is weighing factor for center loss. 
% Next, for every category $y$ present in a mini-batch, the corresponding center deviation $\Delta \mathbf{c}_y$ is used to update $\mathbf{c}_y$, during training.
Consequently, the intra-class and inter-domain variances for each class get minimized, resulting in a structured and domain-agnostic latent space. Furthermore, since both the visual and semantic latent representations of a class are clustered together, the latent space is partitioned based on class-level semantic concepts.

% Impact of our structured partitioning for the DG task on PACS~\cite{basicdg4}. Partitioning the latent space according to the class-level semantic concepts along with multimodal alignment provides notable performance gains (blue bars), in comparison to only aligning the multimodal visual and semantic cues (orange bars). The proposed latent space partitioning achieves significant gains especially on the most difficult unseen domain, \ie, \textit{sketch}.

In order to  validate our hypotheses that a domain-agnostic structured latent space helps to stabilize $p(y|\mathbf{z})$ and generalize to new domains, we conduct an experiment as a proof of concept.  Fig.~\ref{fig:dg} presents a comparison for the standard domain generalization (DG) setting on the PACS dataset~\cite{basicdg4} using ResNet-18 backbone. 
%Impact of our proposed structured partitioning of the latent space for the domain generalization task on the PACS dataset.
%Partitioning the latent space according to the class-level semantic concepts along with multimodal alignment provides notable performance gains (blue bars), in comparison to only aligning the multimodal visual and semantic cues (orange bars). Specifically, the proposed latent space partitioning achieves significant gains on the most difficult unseen domain, \ie, \textit{sketch}.
% The orange bars show performance (domain-wise and average) of our method while employing the multimodal alignment module alone and the blue bars represent the performance while the latent space is structured according to class-level semantic concepts along with multimodal alignment. 
We see that structuring the latent space (blue bars) provides performance gains on all domains and enhances the average gain, compared to employing multimodal alignment alone (orange bars). The highest gain is achieved for the most difficult \emph{sketch} domain that has a large domain shift from the source domains (\emph{photo, art, cartoon}), demonstrating the advantage of our domain-agnostic partitioning.
 
% \begin{SCfigure}
%   \caption{Impact of our structured partitioning of the latent space for the domain generalization task on the PACS dataset. Partitioning the latent space according to the class-level semantic concepts along with multimodal alignment provides notable performance gains (blue bars), in comparison to only aligning the multimodal visual and semantic cues (orange bars). The proposed latent space partitioning achieves significant gains espicially on the most difficult unseen domain, \ie, \textit{sketch}.}
%  \includegraphics[width=0.45\textwidth]{bar_plot.png}
  
% \end{SCfigure}

% \vspace{-3pt}
\vspace{-7pt}
\subsection{Joint Invariance Module\label{triplegan}}
\vspace{-3pt}
As discussed above, the multimodal alignment and conceptual partitioning result in a structured and domain-agnostic latent embedding space that disentangles semantic and domain-specific information.
Such a disentanglement of semantic and domain-specific information is sufficient for standard domain-generalization setting where images during training and testing come from same categories.  However in our ZSLDG setting,  the
disentanglement may not hold for unseen semantic categories during testing, as previously found in \cite{cumix}. 
In order to address this issue and enable generalization to unseen classes in unseen domains, we propose to learn the domain-invariance \wrt{} the joint distribution of visual and semantic representations of a class.
%This might be sufficient for standard domain-generalization setting where images during training and testing come from same categories and we can rely only on disentangling semantic and domain-specific information.
%However in our ZSLDG setting, there is no guarantee that thedisentanglement will hold for the unseen semantic categories at test time \cite{cumix}.
%In order to address this issue and enable generalization to unseen classes in unseen domains, we propose to learn the domain invariance \wrt{} the joint distribution of visual and semantic representations of a class.
% , \ie, $p(f(\mathbf{x}),\mathbf{a}_y)$.\\
Formally, any given image $\mathbf{x}$ comprises of a class-specific content $\textbf{C}$ and a domain-specific transformation $T(\cdot)$ which depicts the class in that particular domain $d$. Thus, each image $\mathbf{x} \in \mathcal{X}$ belonging to domain $d_{i}$ can be represented as  $\mathbf{x} = T_{i}(\textbf{C})$.
% \begin{align}
% \label{eq:transform}
%   x = T_{i}(C),
% \end{align}
% where $T_{i}(.)$ corresponds to the transformation required to depict class with content $C$, in domain $d_{i}$.
% Note that the content (i.e $C$) describes the class-level (domain-invariant) visual cues of class $y$ and thus is correlated to the semantic attributes $a_{y}$ of that particular class. 
In order to enable generalization to unseen class in unseen domains, we propose to match the visual-semantic joint distribution $p(T(\textbf{C}),\mathbf{a}_y)$ under different transformations  $T_{i}(\cdot)$ (or domains). Since the semantic space is shared between seen and unseen classes, learning domain invariance \wrt{} the joint distribution of visual and semantic representations of a class, \ie, $p(f(T(\textbf{C})),\mathbf{a}_y)$ or $p(f(\mathbf{x}),\mathbf{a}_y)$ enables us to enhance generalization.

% However, since the domain invariance is enforced through the marginal distribution $p(f(\mathbf{x}))$, it is likely to overfit to the seen domains with sub-optimal generalization on the target unseen domains~\cite{dgoverfit,condinvariant}. Furthermore, a standard adversarial alignment (Eq.~\ref{eq:align}) is known to suffer from mode collapse~\cite{ganmodes}, as in~\cite{clswgan,taco}, where the model is unlikely to capture the different modes of a complex data distribution. Consequently, the generalization to novel classes and domains at test time decreases. 

% To overcome the aforementioned issues, we propose to learn the domain invariance \wrt{} the joint distribution of visual and semantic representations of a class, \ie, $p(f(\mathbf{x}),\mathbf{a}_y)$ and enhance the visual-semantic space interactions for efficient knowledge transfer. 
Specifically, we aim to match the visual-semantic joint distribution from the visual encoder  $(\mathbf{z}_{v},\mathbf{a}_{y})$, semantic encoder $(\mathbf{z}_{a},\mathbf{a}_{y})$ and projection classifier $(\mathbf{z}_{a},\hat{\mathbf{a}}_{y})$. To this end, we employ a triple adversarial loss, to stabilize the visual-semantic joint distribution across different domains. This also enhances visual-semantic interaction for learning class-specific discriminative features in the visual and semantic embedding spaces. This is achieved by employing a discriminator $D_2: \mathcal{Z} \times \mathcal{A} \rightarrow \mathbb{R}$ and optimizing:
\begin{multline}
\label{eq:tripleD2}
\mathcal{L}_{D_2} =\mathbb{E}[D_2(\mathbf{z}_{a}, \mathbf{a}_y)] - \alpha\mathbb{E}[D_2({\mathbf{z}_{a}}, \hat{\mathbf{a}}_y)]- \beta\mathbb{E}[D_2({\mathbf{z}_{v}}, \mathbf{a}_y)] \\ -  \lambda \mathbb{E}[\left(||\nabla_{\tilde{z}} D_2( \tilde{\mathbf{z}},\tilde{\mathbf{a}}_y),\nabla_{\tilde{\mathbf{a}}} D_2( \tilde{\mathbf{z}},\tilde{\mathbf{a}}_y)||_2 - 1\right)^2]
% \vspace{-3pt}                 
\end{multline}
Here, $\hat{\mathbf{a}}_y=h(\mathbf{z}_a)$ is output from projection classifier $h$, which represents the projection of the latent embedding $\mathbf{z}_a$ onto the semantic space $\mathcal{A}$. Also,
$\tilde{\mathbf{z}} = \eta \mathbf{z}_{a} +(1-\eta)(\alpha\mathbf{z}_a+\beta\mathbf{z}_v)$ and $\tilde{\mathbf{a}}_y = \eta \mathbf{a}_y +(1-\eta)(\alpha\hat{\mathbf{a}}_y+\beta\mathbf{a}_y)$ with $\beta=1-\alpha$ and $\eta \sim U(0, 1)$. Additionally, $\lambda$ is a weighting coefficient. Note that $D_{2}$ is different from the vanilla discriminator $D_{1}$ and has a triple adversarial formulation~\cite{triplegan}. Firstly, by incorporating the projection classifier output $\hat{\mathbf{a}}_y$, it enables to jointly train the visual encoder $f$, semantic encoder $g$ and projection classifier $h$ while imposing domain-invariance. In addition, we design Eq.~\ref{eq:tripleD2} to treat $(\mathbf{z}_{a},\mathbf{a}_{y})$ as real samples and $(\mathbf{z}_{v},\mathbf{a}_{y})$ , $(\mathbf{z}_{a},\hat{\mathbf{a}}_{y})$ as fake samples. This acts as a minimizer of the reverse KL divergence \ie, $KL[(z_{a},a_{y})||(z_{v},a_{y})]$ ~\cite{dualdisc} (in contrast to $D_{1}$ that minimizes forward KL as described in Sec. \ref{alignment}) between the visual and semantic spaces. We find that this leads to better generalization by alleviating the mode collapse issue, and thus enables our model to capture multiple modes of the data distribution~\cite{dualdisc}. Next, the semantic projector classifier $h$ is updated to minimize:
\vspace{-9pt}
\begin{align}
\label{eq:semproj}
\mathcal{L}_{cls} = -\alpha\mathbb E[p_h(y | \mathbf{z}_{a})D_2(\mathbf{z}_{a}, \hat{\mathbf{a}}_y)] + \gamma[ \mathcal{L}_{V}(\mathbf{z}_v,\mathbf{a}_y) + \mathcal{L}_{S}(\mathbf{z}_a,\mathbf{a}_y)], 
% \vspace{-3pt}                 
\end{align}
where $p_h(y|\mathbf{z}_{a})$ is the probability distribution after taking softmax of semantic projection classifier $h$, output logits. Weighting the $D_2$ output with the class probabilities helps in achieving stable training \cite{triplegan}. 
% \highlight{How is it stable training? Some lines are required for better understanding}.  
Finally, we update the visual and semantic encoders ($f$ and $g$) to minimize discrepancy between the embeddings ($\mathbf{z}_{a}$ and $\mathbf{z}_{v}$) in the latent space, given by:
\vspace{-6pt}
\begin{align}
\label{eq:genloss}
\mathcal{L}_{gen} = \mathbb E[D_2(\mathbf{z}_{a}, \mathbf{a}_y)] - \beta\mathbb E[D_2({\mathbf{z}_{v}}, \mathbf{a}_y)].
% \vspace{-3pt}                 
\end{align}
% 
% 
% Then, the joint invariance loss term $\mathcal{L}_{joint-inv}$, which combines the $\mathcal{L}_{cls}$ and $\mathcal{L}_{gen}$ loss terms is defined as $\mathcal{L}_{joint-inv} = \mathcal{L}_{cls} + \mathcal{L}_{gen}$.
% Next, we combine the loss terms $\mathcal{L}_{cls}$ and $\mathcal{L}_{gen}$  into a single loss and define,
% 
% 
% 
Then, the joint invariance loss term $\mathcal{L}_{joint-inv}$ is defined as $\mathcal{L}_{joint-inv} = \mathcal{L}_{cls} + \mathcal{L}_{gen}$.
Consequently, the adversarial loss terms in Eq.~\ref{eq:tripleD2} and $\mathcal{L}_{joint-inv}$ together enable us to jointly train $f,g,h$ and learn a domain-invariant space, which can generalize to unseen domains and classes at test time, by capturing class-specific discriminative visual-semantic relationships across domains.
% \begin{equation}
%     \mathcal{L}_{total} = \mathcal{L}_{align} + \mathcal{L}_{center} + \mathcal{L}_{joint-inv}.
% \end{equation}

\vspace{-6pt}
\subsection{Training and Inference\label{train_test}}
\vspace{-3pt}
% Here, we describe our training and inference procedures of our proposed method. Each of the functions learned in our proposed framework, \ie, $f, g, h, D_{1}, D_{2}$ are parameterized with parameters $\theta_{f}, \theta_{g}, \theta_{h}, \theta_{D_{1}}, \theta_{D_{2}}$, respectively and learned end-to-end. Furthermore, we denote parameters of the center loss as $\theta_{C} = \{\mathbf{c}_1,\ldots,\mathbf{c}_S\}$.

\noindent\textbf{Training:} In a single training iteration, we first update the discriminators $D_{1}$ and $D_{2}$ to maximize the losses in Eq.~\ref{eq:wganD1} and~\ref{eq:tripleD2}.
% \vspace{-3pt}
% \begin{align}
% \label{eq_inference_loss_function}
% \max_{\theta_{D_{1}},\theta_{D_{2}}} \mathcal{L}_{D_{1}} + \mathcal{L}_{D_2}
% \vspace{-3pt}
% \end{align}
We update the discriminators $5$ times for every update of the rest of the functions ($f, g, h$), as in WGAN~\cite{wgangp}. Following this, the parameters $\theta_{f}, \theta_{g}, \theta_{h}, \theta_{c}$ corresponding to $f, g$, $h$ and class centers, respectively, are updated to minimize:
\vspace{-5pt}
% the loss $\mathcal{L}_{total}$, given by
% 
% 
\begin{equation}
    \mathcal{L}_{total} = \mathcal{L}_{align} + \mathcal{L}_{center} + \mathcal{L}_{joint-inv}.
\end{equation}
% 
% 
% This procedure is repeated at every iteration using an alternating training strategy \cite{ganian}.\\
\noindent\textbf{Inference:} A test image $\mathbf{x}_t$ from a unseen domain and class (in $\mathcal{D}^u$ and $\mathcal{Y}^u$) is projected by encoder $f$ to obtain the corresponding latent embedding $\mathbf{z}_t=f(\mathbf{x}_t)$. The semantic projection classifier $h$ computes pairwise similarities between $\mathbf{z}_t$ and the unseen class embeddings $\mathbf{a}_y$, where $y\in\mathcal{Y}^u$. These similarity scores are converted to class probabilities to obtain the final prediction $\hat{y}$, given by
% \begin{equation}
   $ \hat{y} = \argmax_{y\in\mathcal{Y}^u}P(y|\mathbf{x}_{t};\Phi).$
% \end{equation}

\vspace{-7pt}
\section{Experiments\label{sec_expt_results}}
\vspace{-4pt}
\noindent\textbf{Datasets:} We evaluate our method on the DomainNet and DomainNet-LS benchmarks for the task of ZSLDG, as in~\cite{cumix}. \textbf{DomainNet~\cite{domainnet}:} 
% Recently, \cite{cumix} proposed the challenging DomainNet dataset~\cite{domainnet} as a benchmark for the task of zero-shot domain generalization. 
It is a large-scale dataset  and is currently the only benchmark dataset for the ZSLDG setting \cite{cumix}. It consists of nearly $0.6$ million images from $345$ categories in $6$ domains: \textit{painting}, \textit{clipart}, \textit{sketch}, \textit{infograph}, \textit{quickdraw} and \textit{real}.  For the task of ZSLDG, we follow the same training/validation/testing splits along with the training and evaluation protocol described in~\cite{cumix}. In particular, $45$ out of $345$ are fixed as unseen classes and training is performed using only the remaining seen class images. Among the 6 domains in DomainNet, the seen class images from 5 domains are provided during training, and the model is evaluated on the $45$ unseen classes in the held-out (unseen) domain. We repeat experiments with each of the domains as the unseen domain. Following \cite{cumix}, the \textit{real} domain is never held out since a ResNet-50 backbone, pre-trained on ImageNet~\cite{imagenet}, is employed. Average per-class accuracy is used as the performance metric for evaluation on the held-out domain. Similarly, we use the \textit{word2vec}~\cite{word2vec} representations as the semantic information for inter-relating seen and unseen classes, as in \cite{cumix}.\\
\noindent\textbf{DomainNet-LS:} This benchmark is a more challenging setting, where the source domains during training are limited to \emph{real} and \emph{painting} only, whereas testing is conducted on the remaining four unseen domains. Since only two source domains are used in training, it is more challenging to learn domain-invariance and generalize at test-time.

\begin{table*}[t]

\caption{\footnotesize State-of-the-art comparison for the task of ZSLDG on the DomainNet benchmark using ResNet-50 backbone~\cite{cumix}. For a fair comparison, all reported results employ the same backbone, protocol and splits, as described in \cite{cumix}. Best results are in bold.\vspace{0.2cm}}
\centering
\setlength{\tabcolsep}{16pt}
\adjustbox{width=0.9\textwidth}{
\begin{tabular}{c|c|c|ccccc} 
		\toprule[0.15em]
		\multicolumn{2}{c |}{\textbf{Method}} & \textbf{}  & \multicolumn{5}{c }{\textbf{Target Domain}} \\
		\textbf{DG} & \textbf{ZSL} & \textbf{AVG}  & \textit{painting} & \textit{infograph}  & \textit{quickdraw} & \textit{sketch}  &  \textit{clipart}\\
		\midrule

% \begin{tabular}{c|c|c|ccccc}
% \hline
% \textbackslash{}textbf\{DG\} & \textbackslash{}textbf\{ZSL\} & \textbackslash{}textbf\{AVG\} & \textbackslash{}textit\{clipart\} & \textbackslash{}textit\{infograph\} & \textbackslash{}textit\{painting\} & \textbackslash{}textit\{quickdraw\} & \textbackslash{}textit\{sketch\} \\
       %&DEVISE & 26.8 & 12.9  & 23.3  &10.0  & 35.9 & 24.4  &22.7\\
                             %&ALE    & 28.7  & 19.1  & 26.0  & 11.6  & 39.3  & 26.5  & 25.2\\
                      \multirow{3}{*}{-}&       \texttt{DEVISE}~\cite{frome2013devise}    &14.4  &17.6 &11.7   &6.1   &16.7 & 20.1 \\
                     &        \texttt{ALE}~\cite{akata2013label}    &16.2 &20.2 &12.7    &6.8  &18.5 & 22.7   \\
                      &       \texttt{SPNet}~\cite{xian2019semantic}    &19.4  & 23.8   &16.9   & 8.2  & 21.8 &{26.0}  \\
                             
                             		\midrule
		
                             \multirow{3}{*}{\texttt{DANN}~\cite{ganin2016domain}} & \texttt{DEVISE}~\cite{frome2013devise}    &13.9 &16.4 &10.4&7.1&15.1 & 20.5\\
                             
                             & \texttt{ALE}~\cite{akata2013label}    &15.7&19.7 &12.5&7.4&17.9& 21.2 \\
         & \texttt{SPNet}~\cite{xian2019semantic}&19.1 &24.1 &15.8&8.4&21.3 &25.9\\
% 		\cmidrule(lr){2-2}
        \midrule
		
                             \multirow{3}{*}{\texttt{EpiFCR}~\cite{metadg2}} & \texttt{DEVISE}~\cite{frome2013devise} &15.9 &19.3  & 13.9   &7.3  &17.2  & 21.6  \\
                             
                             & \texttt{ALE}~\cite{akata2013label}  &17.5  &21.4 &  14.1  &7.8  &20.9  & 23.2   \\
       & \texttt{SPNet}~\cite{xian2019semantic} %&DEVISE &  &  &  &  &  &  &\\
                             %&ALE    &  &  &  &  &  &  &\\
                               &20.0 & 24.6   & 16.7   & 9.2 & {23.2}  &{26.4}  \\
        % \midrule
        % \multicolumn{2}{|c|}{f-clsWGAN \cite{zslgen0}}&15.1 &20.5 &13.3 &6.6 &14.9 &20.0\\
        % \multicolumn{2}{|c|}{AGG + f-clsWGAN}&21.0 &25.9 &17.0 &11.0 &23.8 &27.4\\ 
        \midrule
        \multicolumn{2}{c|}{CuMix(Mixup-img-only)} & 19.2 & 24.4 & 16.3 & 8.7 & 21.7 & 25.2\\
        \multicolumn{2}{c|}{CuMix(Mixup-two-level)} & 19.9 & 25.3  & 17 & 8.8 & 21.9  & 26.6\\
        \multicolumn{2}{c |}{\texttt{CuMix}~\cite{cumix}}    
                                     & {20.7} & {25.5}  &{17.8}    & {9.9}  &{22.6} &{27.6}   \\
        \midrule
        
        \multicolumn{2}{c |}{\textbf{\texttt{Ours}}}    
                                     & \textbf{21.9}  & \textbf{26.6} &\textbf{18.4}   & \textbf{11.5}  &\textbf{25.0}   &\textbf{27.8}  \\

        \bottomrule[0.15em]
		\end{tabular}
		}
		
		\label{tab:domainnet-additional}
		\vspace{-0.2cm}
\end{table*}
\begin{table}[t]\footnotesize
\begin{minipage}{0.49\linewidth}

\caption{\footnotesize Results on DomainNet-LS with only \textit{real} and \textit{painting} as source domains and ResNet-50 backbone, following protocol described in \cite{cumix}. Best results in bold.\vspace{0.1cm}}
\resizebox{\textwidth}{!}{%
\begin{tabular}{l|c|ccccc}
    \toprule[0.15em]
    \textbf{Model}   & \textbf{AVG} & \textit{quickdraw} & \textit{sketch}  & \textit{infograph} &\textit{clipart} \\
    \midrule
    \texttt{SPNet}& 14.4& 4.8& 17.3& 14.1& 21.5\\ 
    \texttt{Epi-FCR+SPNet}& 15.4 & 5.6& 18.7  & 14.9 & 22.5\\
    \hline
    \texttt{CuMix(MixUp-img-only)}:& 14.3& 4.8 & 17.3& 14.0 & 21.2\\
    \texttt{CuMix(MixUp-two-level)}:& 15.8& 4.9& 19.1 & 16.5  & 22.7\\
    \texttt{CuMix (reverse)}:& 15.4& 4.8& 18.2& 15.8  & 22.9\\
    \texttt{CuMix}:& {16.5} & {5.5}& {19.7} & \textbf{17.1}& {23.7}\\
    \hline
 \texttt{Ours} & \textbf{16.9} &\textbf{7.2} & \textbf{20.5}& 16 & \textbf{24} \\
    \bottomrule[0.15em]
    \end{tabular}}
    
    \label{tab:harderzsldg}
\end{minipage}\hfill
\begin{minipage}{0.49\linewidth}

\caption{\footnotesize Ablation study for different components of our framework on DomainNet dataset for ZSLDG setting. Best results are in bold.\vspace{0.1cm}}
    \centering

\resizebox{\textwidth}{!}{%
\begin{tabular}{l|c|ccccc}
    \toprule[0.15em]
    \textbf{Model}   & \textbf{AVG} & \textit{painting} & \textit{infograph}  & \textit{quickdraw} & \textit{sketch} &\textit{clipart}\\
    \midrule
    \texttt{M1}: $\mathcal{L}_{align}$ & 18.5 & 22.6  & 16.2  & 9.6 & 20.8 & 23.7   \\
    \texttt{M2}: \texttt{M1} + $\mathcal{L}_{center}$  & 20.5 & 25.4  & 16.9  & 9.8 & 24.0 & 26.4 \\     
    \texttt{M3}: \texttt{M2} + $\mathcal{L}_{joint-inv}$ & \textbf{21.9}  & \textbf{26.6}  & \textbf{18.4}  & \textbf{11.5} & \textbf{25.0} & \textbf{27.8}  \\
    %  \texttt{M1}: $\mathcal{L}_{align}$ & 18.5 & 22.6  & 16.2  & 9.6 & 20.8 & 23.7   \\
    % \texttt{M2}: $\mathcal{L}_{align}$ + $\mathcal{L}_{center}$  & 20.5 & 25.4  & 16.9  & 9.8 & 24.0 & 26.4 \\     
    % \texttt{M3}: $\mathcal{L}_{align}$ + $\mathcal{L}_{center}$ + $\mathcal{L}_{joint-inv}$ & \textbf{21.9}  & \textbf{26.8}  & \textbf{18.4}  & \textbf{11.5} & \textbf{25.0} & \textbf{27.8}  \\
    \bottomrule[0.15em]
    \end{tabular}
}

\label{tab_ablation}
\end{minipage}
\vspace{-0.2cm}
\end{table}
\vspace{-7pt}
\subsection{Results: Comparison with State-of-the-art}
\vspace{-3pt}
\noindent \textbf{Results on DomainNet:} 
Tab.~\ref{tab:domainnet-additional} shows the comparison of our proposed framework with  state-of-the-art methods and all baselines, as established in \cite{cumix}, on the ZSLDG task. 
We first report the performance of standalone ZSL approaches such as \texttt{DEVISE}~\cite{frome2013devise}, \texttt{ALE}~\cite{akata2013label} and \texttt{SPNet}~\cite{xian2019semantic} on the ZSLDG task, followed by the performance achieved by coupling these ZSL approaches with standard DG approaches like \texttt{DANN}~\cite{ganin2016domain} and \texttt{EpiFCR}~\cite{metadg2}.
% %
It is worth noting that coupling the standalone ZSL methods with \texttt{DANN} achieves lower performance than the ZSL method alone in the case of ZSLDG, since standard domain alignment methods have been shown to be ineffective on the DomainNet dataset, leading to negative transfer in some cases~\cite{domainnet}. Furthermore, as noted by~\cite{cumix}, coupling \texttt{EpiFCR} (a standalone DG method) with the standalone ZSL approaches is not straightforward, since it requires careful adaptation that includes re-structuring of the loss terms. In particular, the approach of \texttt{EpiFCR}+\texttt{SPNet} achieves an average accuracy (AVG) of $20.0$ over different target domains.
% %
The recently introduced \texttt{CuMix}~\cite{cumix} approach that targets ZSLDG, employs a curriculum-based mixing policy to generate increasingly complex training samples by mixing up multiple seen domains and categories available during training. The current state-of-art \texttt{CuMix} improves ZSLDG performance over \texttt{EpiFCR}+\texttt{SPNet}, achieving an average accuracy of $20.7$ across the target domains. Our approach outperforms \texttt{CuMix} with an absolute gain of $1.2\%$ average across domains ( $\sim6\%$ relative increase) and achieves average accuracy of $21.9$ across the five target domains, setting a new state of the art. Furthermore, our method achieves consistent gains over \texttt{CuMix} on each of the target domains. \\
\noindent\textbf{Results on DomainNet-LS:} 
Tab.~\ref{tab:harderzsldg} shows the performance comparison on the DomainNet-LS benchmark.
The \texttt{SPNet}~\cite{xian2019semantic} (for standard ZSL) achieves an average accuracy of $14.4$, while its integration with \texttt{EpiFCR}~\cite{metadg2} (a standard DG approach) improves the performance to $15.4$. The current state-of-art \texttt{CuMix}~\cite{cumix} approach for ZSLDG, achieves $16.5$ as the average accuracy across the unseen domains. Despite the limited information available during training (and higher domain shift at test time), our approach improves over \texttt{CuMix} by achieving an average accuracy of $16.9$ ($1.6\%$ relative gain), thereby showing better generalization.

\vspace{-7pt}
\subsection{Ablation Study}
\vspace{-3pt}
% In order to study the effectiveness of the proposed method, we conducted various studies to analyze its performance under different settings.
% Here, we present an ablation study to evaluate the contribution of each loss term.
% \noindent \textbf{Contribution of each loss term:} 
We perform an ablation study to understand the efficacy of each component in our proposed method for the ZSLDG task. Tab.~\ref{tab_ablation} shows the performance gains achieved (on DomainNet~\cite{domainnet}) by integrating one contribution at a time, in our approach, as below:
\begin{compactitem}
\item  The model learned by employing our multimodal alignment loss term $\mathcal{L}_{align}$ alone (detailed in Sec.~\ref{alignment}) is denoted as \texttt{M1}
\item Similarly, \texttt{M2} denotes the model learned by integrating $\mathcal{L}_{align}$ with our loss term $\mathcal{L}_{center}$, which achieves a structured latent space (Sec.~\ref{struct_embed}).
\item \texttt{M3} denotes our overall framework, which is learned by integrating our joint invariance loss term $\mathcal{L}_{joint-inv}$  (Sec.~\ref{triplegan}) with $\mathcal{L}_{align}$ and $\mathcal{L}_{center}$. 
\end{compactitem}
The \texttt{M1} model, which performs multimodal adversarial alignment achieves an average accuracy (denoted as AVG in Tab.~\ref{tab_ablation}) of $18.5$ across the target domains. Learning a structured latent embedding space along with the multimodal alignment enables the \texttt{M2} model to achieve an average gain of $2.0$ over \texttt{M1} on the target domains. We note that the gains in \texttt{M2} due to the integration of $\mathcal{L}_{center}$ with $\mathcal{L}_{align}$ are considerably high on the easier target domains (\textit{clipart}, \textit{painting} and \textit{sketch}). This suggests that $\mathcal{L}_{center}$ is able to achieve an improved structuring of the latent embedding space. Our overall framework (\texttt{M3}) obtains the best results by achieving an average accuracy of $21.9$ on the five target domains. Since \texttt{M3} additionally involves learning the domain invariance \wrt{} the visual-semantic joint by employing $\mathcal{L}_{joint-inv}$,
% along with $\mathcal{L}_{align}$ and $\mathcal{L}_{center}$, 
it aids in improving ZSLDG performances on harder target domains such as \textit{quickdraw} and \textit{infograph}. These results clearly indicate that along with the multimodal alignment ($\mathcal{L}_{align}$), structuring the latent space ($\mathcal{L}_{center}$) and learning the domain invariance \wrt{} the visual-semantic joint ($\mathcal{L}_{joint-inv}$) are important for recognizing unseen classes in unseen domains. 

% \textit{Additional ablations and analysis are presented in the supplementary.}

%This paper address  the challenging  problem of recognizing unseen classes in unseen domains, named zero-shot domain generalization. In this work, we propose a novel method that learns a domain-agnostic structured latent embedding space, where multi-domain visual images and semantic class descriptions are projected.
\vspace{-10pt}
\section{Conclusions}
\label{conclusion}
\vspace{-7pt}
We propose a novel approach to address the challenging problem of recognizing unseen classes in unseen domains (ZSLDG). Our method learns a domain-agnostic structured latent embedding space which is achieved by employing a multimodal alignment loss term that aligns the visual and semantic spaces, a center loss term that separates different classes in the latent space and a joint invariance term that aids in handling new classes from unseen domains. Our experiments and ablation studies on challenging benchmarks (DomainNet, DomainNet-LS) show the superiority of our approach over existing methods. 
Future directions include leveraging self-supervision to obtain domain-invariant features and tackle dynamic changes in the label space of categories.

\bibliography{bmvc_final}
\end{document}